\documentclass{article}
\usepackage{spconf,amsmath,graphicx}
\usepackage{amsfonts}
\usepackage{import}
\usepackage{booktabs}
\usepackage{tablefootnote}
\usepackage{multirow}

\title{stream attention-based multi-array end-to-end speech recognition}
\name{Xiaofei Wang$^{1,\ddagger}$, Ruizhi Li$^{1,\ddagger}$, Sri Harish Mallidi$^2$, Takaaki Hori$^3$, Shinji Watanabe$^1$, Hynek Hermansky$^1$ \thanks{${\ddagger}$ Both authors contributed equally to this work. This work is supported by a Google faculty award to Hynek Hermansky and National Science Foundation under Grant No. 1704170.}}
\address{$^1$The Johns Hopkins University, $^2$Amazon, $^3$Mitsubishi Electric Research Laboratories (MERL)\\
{\small \tt \{xiaofeiwang, ruizhili, shinjiw, hynek\}@jhu.edu, mallidih@amazon.com, thori@merl.com}
}

\begin{document}
\ninept

\maketitle

\begin{abstract}
Automatic Speech Recognition (ASR) using multiple microphone arrays has achieved great success in the far-field robustness.
Taking advantage of all the information that each array shares and contributes is crucial in this task.
Motivated by the advances of joint Connectionist Temporal Classification (CTC)/attention mechanism in the End-to-End (E2E) ASR, a stream attention-based multi-array framework is proposed in this work.
Microphone arrays, acting as information streams, are activated by separate encoders and decoded under the instruction of both CTC and attention networks. In terms of attention, a hierarchical structure is adopted.
On top of the regular attention networks, stream attention is introduced to steer the decoder toward the most informative encoders.
Experiments have been conducted on AMI and DIRHA multi-array corpora using the encoder-decoder architecture. Compared with the best single-array results, the proposed framework has achieved relative Word Error Rates (WERs) reduction of $3.7\%$ and $9.7\%$ in the two datasets, respectively, which is better than conventional strategies as well.

%The framework assigns discriminate attentions for the array streams, resulting in a hierarchical attention on top of the regular attention in the encoder-decoder architecture.
%
\end{abstract}

%

% sec1. Introduction:
% 1. explain the multi-array setting and difficulties, and stress the importance of the information fusion of this setting in terms of robustness, and how different this task is from multi-channel -- xiaofei
% 2. prior work in conventional ASR (multi-array and multi-channel work). Specially, how fusion is done.  (front-end fusion, back-end fusion, posterior fusion, PM) -- xiaofei
% 3. description of end-to-end approach and its advantage. Introduction of joint e2e model.  --ruizhili 
% 4. prior work in e2e (multi-array and multi-channel). specially, how fusion is done(frame-level fusion(multi-head), context vector level fusion(HAN)) --ruizhi
% 5. our introduce (hierarchical attention network) HAN in joint e2e model. (highlight: 1. multi-encoder for multi-array setup 2. HAN for robustness, 2. joint model with per-encoder ctc helps with ctc in general) -- xiaofei

\section{Introduction}
\label{sec:intro}

Far-field ASR using multiple microphone arrays has been a widely adopted strategy in the speech processing community. 
Individually, the microphone array is able to bring a substantial performance improvement with algorithms such as beamforming \cite{vincent2017analysis} and masking \cite{wang2016oracle}. 
However, what kind of information can be extracted from each array and how to make multiple arrays work in cooperation are still challenging. 
Without any prior knowledge of speaker-array distance or video monitoring, it is difficult to figure out which array carries more reliable information or is less corrupted.  

According to the reports from the CHiME-5 challenge \cite{Barker2018}, which targets the problem of multi-array conversational speech recognition in home environments, the common ways of utilizing multiple arrays in the hybrid ASR system are finding the ones with higher Signal-to-Noise/Interference Ratio (SNR/SIR) \cite{du2018theustc} or fusing the decoding results by voting for the most confident words \cite{kanda2018hitachi}, e.g. ROVER \cite{fiscus1997post}.
Similar to our previous work \cite{wang2018stream}, combination using the classifier's posterior probabilities followed by lattice generation has been an alternative approach \cite{xiong2018channel}.
%, which uses higher-level representations to derive the confidence of each array stream.
The posteriors from the well-trained classifier decorrelate the input features, but reserve more distinctive speech information than the words after the full decoding stage.
In terms of the combination strategy, ASR performance monitors have been designed  \cite{mallidi2015uncertainty}, resulting in a process of stream confidence generation, guiding the linear fusion of array streams.
%In this case, extra ASR performance monitors are in demand, which makes the application possible by assigning a linear weight for each stream. Essentially, it is a stream attention mechanism.

Recently, E2E ASR has attracted attention in the research field. The E2E system is developed to directly transcribe human speech into text. 
It integrates disjoint modules, developed from traditional hybrid methods, into one single Deep Neural Network (DNN) which can be trained from scratch.
The attention-based structure \cite{chan2016listen,chorowski2015attention} solves the ASR problem as a sequence mapping by using an encoder-decoder architecture. 
Coupled with a CTC network \cite{graves2014towards,miao2015eesen}, the joint model \cite{kim2016joint_icassp2017,hori2017advances,watanabe2017hybrid} outperforms the attention-based ASR by addressing misalignment issues. 
While most of the E2E ASR studies engage in single-channel task or multi-channel task from one microphone array \cite{ochiai2017unified,braun_2018,ochiai2017multichannel,kim2017end}, research on multi-array scenario is still unexplored within the E2E framework. 
% End-to-End (E2E) ASR has achieved great success using the encoder-decoder structure, in which attention plays an important role in addressing the crucial part of the inputs \cite{kim2017joint}.
% Coupled with a connectionist temporal classification (CTC) network, E2E ASR assembles both the advantages from attention and CTC \cite{kim2016joint_icassp2017,hori2017advances,watanabe2017hybrid}. Not only is the speech-label monotonic property satisfied, but the sequential property is also integrated through introduction of the CTC loss into the loss function for training.

In this work, we propose an attention-based multi-array E2E architecture -- the joint CTC/Attention model with the hierarchical attention mechanism, inspired by our original work \cite{li2018multi} done at JSALT 2018, to solve the aforementioned problem. The framework has highlights as follows: 
\begin{enumerate}
\item The output of each microphone array is modeled by a separate encoder. Multiple encoders with the same configuration act as the acoustic models for individual arrays.
\item The hierarchical attention mechanism \cite{yang2016hierarchical, hori2017attention, libovicky2017attention} was introduced to dynamically combine knowledge from parallel streams. We adopt this network in multi-array scheme, where the stream-level fusion is employed on top of the per-encoder attention mechanisms. 
% \item On top of the per-encoder attention mechanism, hierarchical attention network \cite{yang2016hierarchical, hori2017attention, libovicky2017attention} is employed to steer toward the array stream, who  more informative higher-level representations;
\item Each encoder is associated with a CTC network to guide the frame-wise alignment process for each array to potentially achieve a better performance. 
\end{enumerate}

The remainder of this paper is organized as follows: 
Section 2 reviews previous work.
The proposed multi-stream framework is presented in Section 3, followed by experiments and analysis in Section 4. In the end, the conclusion is given Section 5.

\section{PRIOR WORK}
\label{sec:prior}

\subsection{Conventional Multi-Array ASR}
In our previous work, we proposed a stream attention framework to improve the far-field performance in the hybrid approach, using distributed microphone array(s) \cite{wang2018stream}. 
Specifically, we generated more reliable Hidden Markov Model (HMM) state posterior probabilities by linearly combining the posteriors from each array stream, under the supervision of the ASR performance monitors. 

% Mathematically, given the posterior probability sequences $P_t=[{\bf p}_t^1,{\bf p}_t^2,...,{\bf p}_t^M]^T$ of the HMM states $O$ at time $t$ (where $[*]^T$ is the transpose operation, ${\bf p}_t^i=p(O|X_t^i), i=1,...,M$ is the $i$th posterior probability sequence given the context based feature sequence $X_t^i=[{\bf x}_{t-\tau}^i,...,{\bf x}_t^i,...,{\bf x}_{t+\tau}^i]^T$ extracted from the signal of microphone $i$ and $M$ is the total stream number), the linearly weighted posterior sequence can be obtained by
% \begin{equation}
% \label{equ:combination}
% \hat{{\bf p}_t}={\bf w}_tP_t
% \end{equation}
% This is used for decoding. 
% The weight $w_t^i\in{\bf w}_t$ for each stream was normalized to a scalar between $0$ and $1$ based on the quality measurement of the posteriors from Deep Neural Network (DNN). We proposed using mean time distance \cite{hermansky2013mean} and Time-delayed DNN auto-encoder \cite{wang2018stream} as the measurements.
In general, the posterior combination strategy outperformed conventional methods, such as signal-level fusion and the word-level technique ROVER \cite{fiscus1997post}, in the prescribed multi-array configuration. Accordingly, stream attention weights estimated from the de-correlated intermediate features should be more reliable. We adopt this assumption in the following context.

\subsection{Joint CTC/Attention Architecture for End-to-End ASR}
The joint CTC/Attention model for the E2E ASR outperforms ordinary attention-based ones by solving the misalignment issues between the speech and labels \cite{kim2016joint_icassp2017,hori2017advances,watanabe2017hybrid}. It digests the advantages of both CTC and attention-based model through a multi-task learning mechanism and joint decoding. Accordingly, the E2E model maps $T$-length acoustic features $X=\{\textbf{x}_{t}\in \mathbb{R}^{D}|t = 1,2,...,T\}$ in $D$ dimensional space to an $L$-length letter sequence $C=\{c_{l}\in \mathcal{U}|l = 1,2,...,L\}$ where $\mathcal{U}$ is a set of distinct letters.

The encoder is shared by both attention and CTC networks. Typical Bidirectional Long Short-Term Recurrent (BLSTM) layers are utilized to model the temporal dependencies of the input sequence. The frame-wise hidden vector $\textbf{h}_{t}$ at frame $t$ is derived by encoding the full input sequence $X$:
\begin{equation}
\textbf{h}_{t}=\textrm{Encoder}(X)
\end{equation}

For the attention-based encoder-decoder model, the letter-wise context vector $\textbf{r}_{l}$ is formed as a weighted summation of frame-wise hidden vectors $\textbf{h}_{t}$ using a content-based attention mechanism:
\begin{equation}
\label{f:att1}
\textbf{r}_{l}={\sum}_{t=1}^{T}a_{lt}\textbf{h}_{t},\ {a}_{lt}=\textrm{ContentAttention}(\textbf{q}_{l-1}, \textbf{h}_t)
\end{equation}
where $\textbf{q}_{l-1}$ is the previous decoder state, and ${a}_{lt}$ is the attention weight, a soft-alignment of $\textbf{h}_{t}$ for $c_{l}$. An LSTM-based decoder network predicts the next letter based on $\textbf{r}_{l}$ and the previous prediction.

% For the CTC network, the frame-wise posterior probability $p(z_{t}|X)$ is estimated in the following way:
% \begin{equation}
% p(z_{t}|X)=\textrm{Softmax}(\textrm{LinB}(\textbf{h}_{t})) 
% \end{equation}
% where $\textrm{LinB}(\cdot)$ is a linear layer with bias term converting $\textbf{h}_{t}$ to a ($|\mathcal{U}|+1$) dimensional vector, followed by a Softmax activation function and $z_{t}\in \mathcal{U} \bigcup blank$. Note that the ``blank'' symbol is used to handle the merging of repeating letters.

The objective function to be maximized is as follows:
\begin{equation} 
\label{equ:loss}
\mathcal{L}=\lambda\log p_{ctc}( C|X)+(1-\lambda)\log p_{att}^{\dagger}(C|X)
\end{equation}
where the joint objective is a logarithmic linear combination of the CTC and attention training objectives, i.e., $p_{ctc}(C|X)$ and $p_{att}^{\dagger}(C|X)$, respectively. 
The attention $p_{att}(C|X)$ is approximated during training as $p_{att}^{\dagger}(C|X)$, where the probability of a prediction is conditioned on previous true labels. 
$0\leq\lambda\leq 1$ is a trade-off parameter satisfying.
In the decoding phase, the joint CTC/Attention model performs a label-synchronous beam search which jointly predicts the next character. The most probable letter sequence $\hat C$ given the speech input $X$ is computed as:
\begin{align}
\label{equ:max}
    \hat{C}=\arg\max_{C\in \mathcal{U}^{*}} &\{\lambda \log p_{ctc}(C|X)+(1-\lambda)\log p_{att}(C|X) \nonumber \\
    &+\gamma \log p_{lm}(C)\} 
\end{align}
where an external Recurrent Neural Network Language Model (RNNLM) probability $\log p_{lm}(C)$ is added with a scaling factor $\gamma$.
% For each partial hypothesis $h$ in the beam search, the joint score, the log probability of hypothesized label sequence, can be computed as 
% \begin{equation}
%     \alpha_{att}(h)=\lambda \alpha_{ctc}(h)+(1-\lambda)\alpha_{att}(h)+\gamma\alpha_{lm}(h),
% \end{equation}
% where the attention decoder score, $\alpha_{att}(h)$, can be accumulated recursively from hypothesis scores from one step before. 
% In terms of CTC score, $\alpha_{ctc}(h)$, CTC prefix probability, defined as the cumulative probability of all label sequences that have $h$ as their prefix, was used. In this work, we use the look-ahead word-based language model to give the RNNLM score, $\alpha_{lm}(h)$. This language model decodes with only a word-based model \cite{hori2018end}, avoiding  using a multi-level LM which uses a character-level LM until the identity of the word is determined. 

\section{Multi-Array End-to-End Model}
\label{sec:proposed}
\begin{figure}[htb]
  \centering 
  \centerline{\includegraphics[width=6cm]{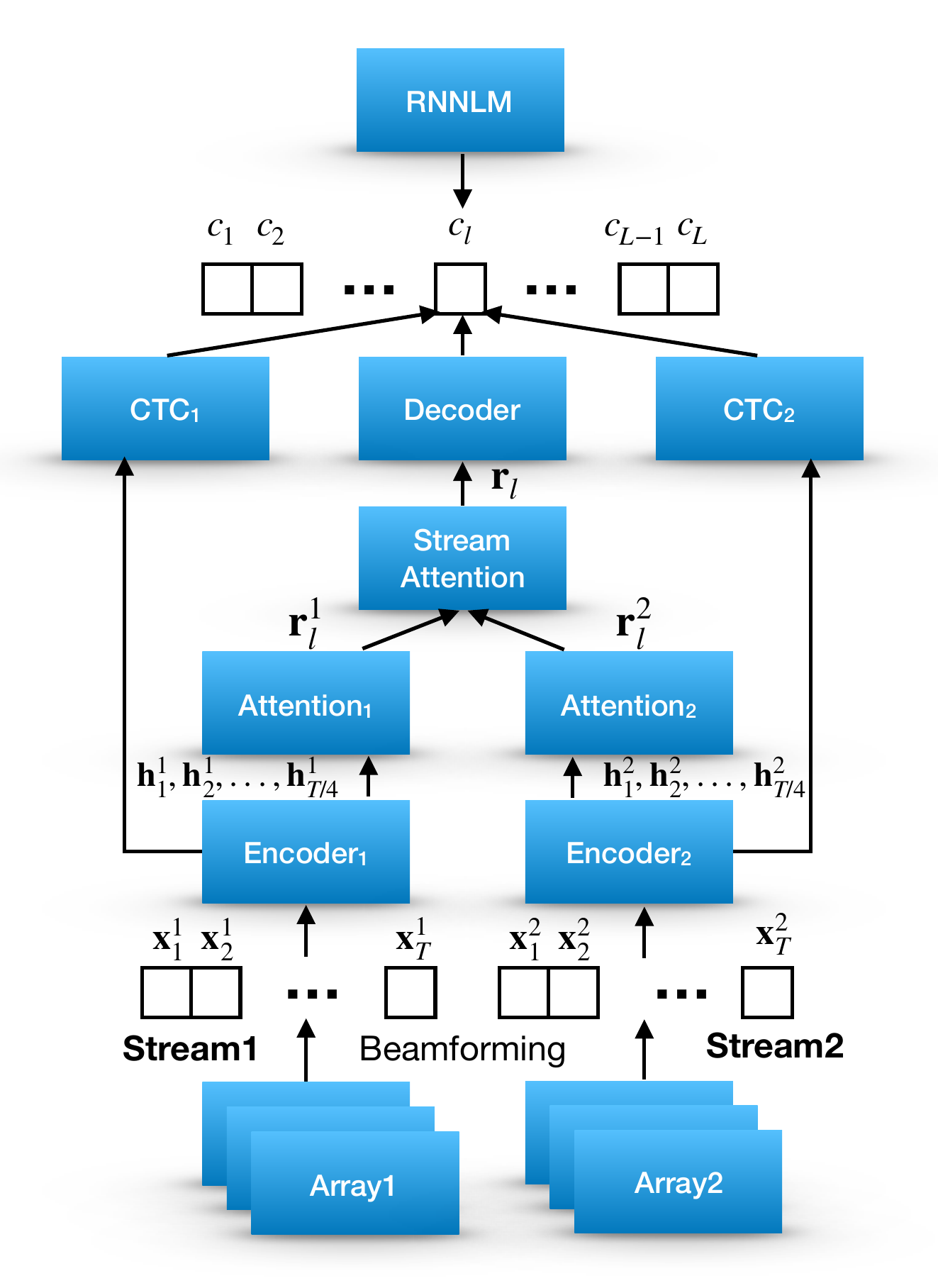}}
\caption{Multi-Stream Architecture Using Two Microphone-Arrays.}
  \label{fig:multi}
\end{figure}

In this section, we present the stream attention based E2E framework \cite{li2018multi} for the multi-array ASR task. A hierarchical attention scheme is introduced within the CTC/Attention joint training and decoding mechanism. For simplicity to understand the framework, we focus on the two-array architecture, which is shown in Fig.\ref{fig:multi}.

\subsection{Multi-Array Architecture with Stream Attention}
% sec3. proposed model (han in joint model framework) 
% 3.1-2. multi-encoder with HAN (we explore two kinds of encoder structure) 
% (, subsampling, importance of two encoders, han fusion allowing input with different time, instead of frame-level fusion, content attention is used here, with p and without p)
% 3.3. joint model with per-encoder CTC (trained on the paralell data but not reuiqred to bethe same time-alingment, explain training and decoding with RNNLM, what kinda rnn-lm is used here, sharectc ctc is good why?, )

\label{sssec:encarch}

The proposed architecture has two encoders, with each mapping the speech features of a single array to higher level representations ${\bf h}_t^i$, where we denote $i\in\{1,2\}$ as the index for ${\textrm{Encoder}_i}$ corresponding to array $i$. 
Note that ${\textrm{Encoder}_1}$ and ${\textrm{Encoder}_2}$ have the same configurations receiving parallel speech data collected from multiple microphone arrays. 
Convolution Neural Networks (CNN) are often used together with BLSTM layers on top to extract frame-wise hidden vectors. 
We explore two types of encoder structures: BLSTM (RNN-based) and VGGBLSTM (CNN-RNN-based) \cite{cho2018multilingual}:
\begin{equation}
\label{f:blstm}
\textbf{h}_{t}^{i}=\textrm{Encoder}_{i}(X), i\in\{1,2\}\\
\end{equation}
\begin{equation}
\textrm{Encoder}_{i}()=\textrm{BLSTM}() \quad or \quad \textrm{VGGBLSTM}()
\end{equation}

Note that the BLSTM encoders are equipped with an additional projection layer after each BLSTM layer. 
In both encoder architectures, subsampling factor $s=4$ is applied to decrease the computational cost. Specially, the convolution layers of the VGGBLSTM encoder downsamples the input features by a factor of 4 so that there is no subsampling in the recurrent layers. 

In the multi-stream setting, one inherent problem is that the contribution of each stream (array) changes dynamically. Specially, when one of the streams takes corrupted audio, the network should be able to pay more attention to other streams for the purpose of robustness. 
Inspired by the advances of linear posterior combination \cite{wang2018stream} and a hierarchical attention fusion \cite{yang2016hierarchical, hori2017attention, libovicky2017attention}, a stream-level fusion on the letter-wise context vector is introduced in this work to achieve the goal of encoder selectivity. The letter-wise context vectors, $\textbf{r}_l^1$ and $\textbf{r}_l^2$, from individual encoders are computed similar to Eq. (\ref{f:att1}): 
\begin{equation}
\label{f:cv1}
\textbf{r}_{l}^{i}={\sum}_{t=1}^{T/4}a_{lt}^i\textbf{h}_{t}^{i}, i\in\{1,2\}
\end{equation}
where the summation is performed from $1$ to $T/4$ due to subsampling. The fusion context vector $\textbf{r}_l$ is obtained as a combination of $\textbf{r}_l^1$ and $\textbf{r}_l^2$ as illustrated:
\begin{equation}
\label{f:han}
\textbf{r}_{l}=\beta_{l1}\textbf{r}_{l}^{1} + \beta_{l2}\textbf{r}_{l}^{2}
\end{equation}
\begin{equation}
\label{f:l2att}
{\beta}_{li}=\textrm{ContentAttention}(\textbf{q}_{l-1}, \textbf{r}_l^i), i=1,2
\end{equation}
The stream-level attention weights $\beta_{l1}$ and $\beta_{l2}$ are estimated according to the feedback from the previous decoder state, $\textbf{q}_{l-1}$, and context vectors, $\textbf{r}_l^1$ and $\textbf{r}_l^2$, from individual encoders.
The fusion context vector is then fed into the decoder to predict the next letter. 

In comparison to fusion on frame-wise hidden vectors ${\bf h}_t^i$, stream-level fusion can deal with temporal misalignment from multiple arrays at the stream level. Furthermore, adding an extra microphone array $j$ could be simply implemented with an additional term $\beta_{lj}\textbf{r}_{l}^{j}$ in Eq.(\ref{f:han}). 

\subsection{Training and Decoding with Per-encoder CTC}
We assign each encoder with a separate CTC network. 
During multi-task training and joint decoding, we follow the similar formulas depicted by Eq. (\ref{equ:loss}) and Eq. (\ref{equ:max}). The only difference is that we have per-encoder CTC objective to compute the loss:
\begin{equation}
\log p_{ctc}(C|X)=\frac{1}{2}\lambda(\log p_{ctc_1}(C|X)+\log p_{ctc_2}(C|X)),
\end{equation}
where the equal weight is assigned to each CTC network.

% assgined ctc for each encoder 
% training and decoding replace the function as pctc 
% we use look ahead language model \cite  

% The CTC prefix score of hypothesized sequence $h$ is: 
% \begin{equation}
%     \alpha_{ctc}(h)=\frac{1}{2}(\alpha_{ctc_1}(h) + \alpha_{ctc_2}(h)),
% \end{equation}
% where equal weight is assigned to each CTC network. 

\section{Experiment and Discussion}
\label{sec:experiment}

\subsection{Dataset (AMI and DIRHA) Description}
\label{amidescrip}
The AMI Meeting Corpus consists of 100 hours of far-field recordings from 3 meeting rooms (Edinburgh, Idiap and TNO Room) \cite{carletta2005ami}. The recordings use a range of signals synchronized to a common time line. 
There are two arrays placed in each meeting room to record the sentences, with one 10 cm radius circular array between the speakers consisting of 8 omni-directional microphones.
The setups of the second microphone array are different among the rooms, detailed by Table \ref{tab:table1}. 
%The second Edinburgh array is also a 10cm radius circular array with 4 elements placed at the end of the table, while the second Idiap circular microphone array has only four elements and is mounted on the ceiling rather than on the table and the second TNO array is a 10-element linear array mounted above the presentation screen.
%\subsubsection{DIRHA dataset}

The DIRHA dataset was collected in a real apartment setting with typical domestic background noise and reverberation \cite{ravanelli2016realistic}. In the configuration, a total of 32 microphones were placed in the living-room (26 microphones) and in the kitchen (6 microphones). The microphone network consists of 2 circular arrays of 6 microphones (located on the ceiling of the living-room and the kitchen), a linear array of 11 sensors (located in the living-room) and 9 microphones distributed on the living-room walls. During the recording, the speaker was asked to move to a different position and take a different orientation after reading several sentences.

In both datasets, we chose two microphone arrays as parallel streams (noted by Str1 and Str2) to train the proposed E2E system, which is also shown by Table \ref{tab:table1}. For each microphone array, all the simulations or recordings were synthesized into the single channel using delay-and-sum (DS) beamforming with the BeamformIT Toolkit \cite{anguera2007acoustic}. 
The AMI training set consists of 81 hours of speech. The development (Dev) and evaluation (Eval) set respectively contain 9 hours of meeting recordings. We used Dev set for cross validation and Eval set for testing.
Contaminated version of the original WSJ (Wall Street Journal) corpus is used for DIRHA training. Two streams were generated using the WSJ0 and WSJ1 clean utterances convolved by the circular array impulse responses and the linear ones, respectively. Recorded noises were added as well. We used the DIRHA Simulation set (generated via the same way as training data) for cross validation and DIRHA Real set for testing, which consisted of 3 Male and 3 Female native US speakers uttering 409 WSJ sentences.
\begin{table}[htb]
  \begin{center}
   	\caption{Description of the array configuration in the two-stream E2E experiments.}
    \label{tab:table1}
	\begin{tabular}{l|c|c}
	  \toprule
	  Dataset   & Str1 (Stream 1)  & Str2 (Stream 2) \\
	  \midrule
             &                      & Edinburgh: 8-mic Circular Array \\
      AMI    & 8-mic Circular Array & Idiap: 4-mic Circular Array \\
             &                      & TNO: 10-mic Linear Array \\
      \midrule
      DIRHA  & 6-mic Circular Array & 11-mic Linear Array \\
      \bottomrule
	\end{tabular}
  \end{center}
\end{table}

All the experiments were implemented by ESPnet, an end-to-end speech processing toolkit \cite{watanabe2018espnet} with the configuration as described in Table \ref{tab:config}:
\begin{table}[th]
  \begin{center}
   	\caption{Experimental configuration}
    \label{tab:config}
	\begin{tabular}{ll}
	  \toprule
	  {\scriptsize{}{\bf Feature}} \\
% 	  \hline
      {\scriptsize{}Single Stream} & {\scriptsize{}80-dim fbank + 3-dim pitch}\\
      {\scriptsize{}Multi Stream} & {\scriptsize{}$\text{Array}_1$:80+3; $\text{Array}_2$:80+3}\\
      \hline
	  {\scriptsize{}{\bf Model}} \\
% 	  \hline
      {\scriptsize{}Encoder type} & {\scriptsize{}BLSTM or VGGBLSTM}\\
      {\scriptsize{}Encoder layers} & {\scriptsize{}BLSTM:4; VGGBLSTM\cite{cho2018multilingual}:6(CNN)+4}\\
      {\scriptsize{}Encoder units } & {\scriptsize{}320 cells (BLSTM layers)}\\
      {\scriptsize{}(Stream) Attention} & {\scriptsize{}Content-based}\\
    %   {\scriptsize{}Decoder type} & {\scriptsize{}LSTM}\\
      {\scriptsize{}Decoder type} & {\scriptsize{}1-layer 300-cell LSTM}\\
    %   {\scriptsize{}Decoder units} & {\scriptsize{}300}\\
    %   \hline 
% 	  {\scriptsize{}{\bf Train and Decode}} \\
	%  {\scriptsize{}Batch size} & {\scriptsize{}15}\\
	%  {\scriptsize{}Optimizer} & {\scriptsize{}AdaDelta}\\
    %   \hline 
    %   Decoding \\
      {\scriptsize{}CTC weight $\lambda$ (train)} & {\scriptsize{}AMI:0.5; DIRHA:0.2}\\
    {\scriptsize{}CTC weight $\lambda$ (decode)} & {\scriptsize{}AMI:0.3; DIRHA:0.3}\\
        \hline 
        {\scriptsize{}{\bf RNN-LM}} \\
      {\scriptsize{}Type} & {\scriptsize{}Look-ahead Word-level RNNLM \cite{hori2018end}}\\
    %   {\scriptsize{}Train data (AMI)} & {\scriptsize{}AMI}\\
      {\scriptsize{}Train data} & {\scriptsize{}AMI:AMI; DIRHA:WSJ0-1+extra WSJ text data}\\
    %   {\scriptsize{}Word look-ahead RNNLM \cite{hori2018end}} & {\scriptsize{}AMI:AMI data; DIRHA:WSJ0-1\&extra test data}\\
      {\scriptsize{}LM weight $\gamma$} & {\scriptsize{}AMI:0.5; DIRHA:1.0}\\
      \bottomrule
	\end{tabular}
  \end{center}
\end{table}
%The Bi-RNN models mentioned above uses a LSTM cell followed by a projection layer (BLSTMP). In our experiments below, we use a character-level seq2seq model trained by CTC and attention decoder. 
%We integrated a character-level RNNLM with seq2seq model externally and showcase the performance in terms of word error rate (% WER). In this case the words are obtained by concatenating the characters and the space together for scoring with reference words. 
\subsection{Results}
We define two kinds of E2E architectures in these results discussions: single-stream architecture, which has only one encoder without stream attention and multi-stream architecture, which has several encoders with each corresponding to one microphone array and has stream attention mechanism as well.

\subsubsection{Single-array results}
First of all, we explore the ASR performance for the individual array (single stream). 
As illustrated in Table \ref{tab:table3}, the single stream system with the VGGBLSTM based encoder outperforms the one with BLSTM encoder, both in Character Error Rate (CER) and WER. 
Joint training of CTC and attention based model helps since CTC can enforce the monotonic behavior of attention alignments, rather than merely estimating the desired alignment for long sequence.
With the RNNLM, we can see a dramatical decrease of the WERs on both datasets. The Str1 WERs of AMI Eval and DIRHA Real are 56.9\% and 35.1\%, respectively.
For simplicity, we only keep the CTC/Attention based single-stream results with RNNLM for Str2 since the same trend can be found and only the WER will be compared in the following results. 
%Actually, given an utterance in the test set, we have no idea which one is more reliable as the speaker may move around during speaking, even though Str1 seems always better than Str2, overall.
\begin{table}[htb]
  \begin{center}
  	\caption{Exploration of best encoder and decoding strategy for single-stream E2E model.}
    \label{tab:table3}
	\begin{tabular}{l|c|c|c|c}
	  \toprule
	  &\multicolumn{2}{|c|}{AMI} & \multicolumn{2}{c}{DIRHA}  \\
	  Model (Single Stream) & \multicolumn{2}{c|}{Eval} & \multicolumn{2}{c}{Real}  \\
	        & CER & WER &  CER & WER \\
	  \midrule
      {\it BLSTM} (Str1)\ \ \  & & & \\
      Attention  & 45.1 & 60.9 & 42.7 & 68.7 \\
      + CTC      & 41.7	& 63.0 & 38.5 & 74.8\\
      + Word RNNLM   & 41.7	& 59.1 & 29.4 & 47.4\\
      \midrule
      {\it VGGBLSTM} (Str1)\ \ \  & & & \\
      Attention  & 43.2	& 59.7 & 39.5 &	71.4 \\
      + CTC      & 40.2	& 62.0 & 30.1 & 61.8\\
      + Word RNNLM   & \bf 39.6	& \bf 56.9 & \bf 21.2 & \bf 35.1\\
      \midrule
      \midrule
      {\it VGGBLSTM} (Str2)\ \ \  & \bf 45.6	& \bf 64.0 & \bf 22.5 & \bf 38.4 \\
      \bottomrule
	\end{tabular}
  \end{center}
\end{table}

\subsubsection{Multi-array results}
As shown in Table \ref{tab:table4}, the proposed stream attention framework achieves 3.7\% (56.9 to 54.9) and 9.7\% (35.1 to 31.7) relative WERs reduction on AMI and DIRHA datasets, respectively. Hierarchical attention plays a role that emphasizing the more reliable stream.
In addition, we compare the multi-stream framework with conventional strategies using single-stream system trained by the Fbank and pitch features, either concatenated by the Str1 and Str2 features or extracted from the speech audio through alignment and average between the streams. 
The multi-stream framework outperforms the others. 
To explain the improvement is not from the boost of the number of model parameters, we doubled the BLSTM layers (4 to 8) in the VGGBLSTM encoder and train the single-stream CTC/Attention system with a comparable amount of parameters (33.7M vs 31.6M). Our system still has strong competitiveness.
\begin{table}[htb]
  \begin{center}
   	\caption{WER(\%) Comparison between the proposed multi-stream approach and alternative single-stream strategies.}
    \label{tab:table4}
	\begin{tabular}{l|c|c|c}
	  \toprule
	  Encoder {\it VGGBLSTM}  & \#Param & AMI & DIRHA \\
	  (Att + CTC + RNNLM)      &    & Eval  & Real \\
	  \midrule
	  Single-stream model & &\\
	  Concatenating Str1\&Str2 & 23.3M & 56.7 & 33.5 \\
%	  \midrule
%	  WAV alignment and sum  & 16.4M & 58.4 & 43.4 \\
	  WAV alignment and average & 26.2M & 56.7 & 43.5 \\
          + model parameter extension  & 33.7M & 56.9 & 39.6 \\
	  \midrule
	  \midrule
      Multi-stream model  & &   \\
      Proposed framework & 31.6M &  \bf 54.9 & \bf 31.7 \\
      \bottomrule
	\end{tabular}
  \end{center}
\end{table}

\begin{figure}[tb]
  \centering 
  \centerline{\includegraphics[width=8.5cm]{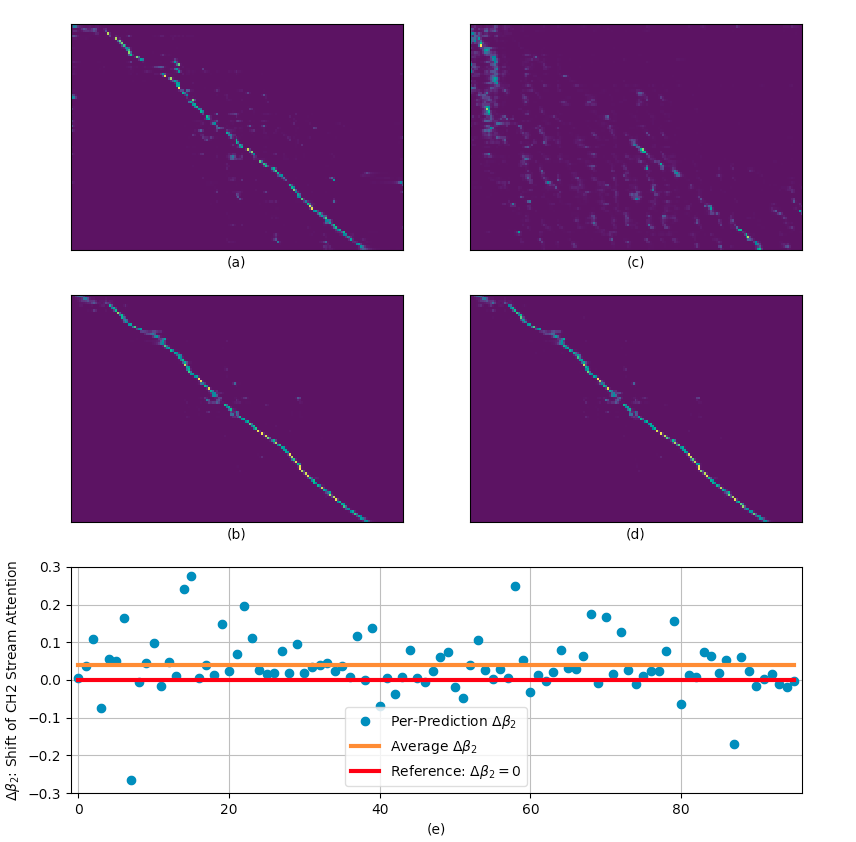}}
\caption{Comparison of the alignments between characters (y-axix) and acoustic frames (x-axis) before (({\bf a}) Str1; ({\bf b}) Str2) and after (({\bf c}) Str1; ({\bf d}) Str2) noise corruption of Str1. ({\bf e}) shows the attention weight shift of Str2 between two cases (x-axis is the letter sequence).}
  \label{fig:corr}
\end{figure}

During the inference stage of the multi-stream model, we examine how the stream attention weights change once one of the streams is corrupted by noise. Fig.\ref{fig:corr} shows an example in the DIRHA Real set that whether the input features of Str1 is affected by an additive Gaussian noise with zero mean and unit variance. After the corruption, the alignment between characters and acoustic frames of Str1 becomes blurred (Fig.\ref{fig:corr}(c)), indicating that the information from Str1 should be less trusted. Therefore, as expected, a positive shift of the attention weights for Str2 can be observed (upper line in Fig.\ref{fig:corr}(e)).

\subsubsection{Comparison with hybrid system}
Table \ref{tab:table5} shows the comparison between the proposed E2E framework and the conventional hybrid ASR approach. In \cite{wang2018stream}, we designed three scenarios using different subsets from the 32 microphones and 2 arrays in the DIRHA dataset. Our proposed DNN posterior combination approach and ROVER technique could relatively reduce the WER of the hybrid system by 7.2\% and 5.8\% respectively, if we average the WERs of the Real test sets among three cases. 
Meanwhile, a relative 9.7\% WER reduction has already been achieved in the stream attention-based two-stream E2E system, even though we have less number of streams (two) than the hybrid one (six). Ignoring the WER gap between the hybrid and E2E ASR systems, we still believe that the proposed E2E approach has much potential to do better with more array streams.
\begin{table}[htb]
  \begin{center}
   	\caption{WER(s) Comparison between the hybrid and end-to-end system on DIRHA dataset. \#Num denotes the number of streams.}
    \label{tab:table5}
	\begin{tabular}{l|c|c|c|c}
	  \toprule
	  System & \#Num & Method & Best Stream & WER  \\
	  \midrule
	  Hybrid   & 6 & post. comb. & 29.2 & 27.1 (\bf 7.2\%) \\
	           & 6 & ROVER & 29.2 & 27.5 (\bf 5.8\%)\\
	  \midrule
	  E2E & 2 & proposed & 35.1 & 31.7 (\bf 9.7\%)\\
      \bottomrule
	\end{tabular}
  \end{center}
\end{table}

\section{Conclusion}
\label{sec:conclusion}
In this paper, we presented a multi-stream End-to-End ASR framework targeting the distributed microphone array situation.
Stream attention was achieved through a hierarchical connection between the decoder and encoders, with each modeling one array into higher-level representations. Thanks to the success of joint training of per-encoder CTC and attention, substantial WER reduction was shown in both AMI and DIRHA corpora, demonstrating the potentials of the proposed architecture. 
For further research, an extension to more streams efficiently and exploration of schedule training of the encoders would be interesting.

%\section{Acknowledgement}
%\label{sec:acknowledgement}

\vfill\pagebreak

% References should be produced using the bibtex program from suitable
% BiBTeX files (here: strings, refs, manuals). The IEEEbib.bst bibliography
% style file from IEEE produces unsorted bibliography list.
% -------------------------------------------------------------------------
\bibliographystyle{IEEEbib}
\bibliography{refs}

\begin{thebibliography}{10}

\bibitem{vincent2017analysis}
Emmanuel Vincent, Shinji Watanabe, Aditya~Arie Nugraha, Jon Barker, and Ricard
  Marxer,
\newblock ``An analysis of environment, microphone and data simulation
  mismatches in robust speech recognition,''
\newblock {\em Computer Speech \& Language}, vol. 46, pp. 535--557, 2017.

\bibitem{wang2016oracle}
Ziteng Wang, Xiaofei Wang, Xu~Li, Qiang Fu, and Yonghong Yan,
\newblock ``Oracle performance investigation of the ideal masks,''
\newblock in {\em IWAENC 2016}. IEEE, 2016, pp. 1--5.

\bibitem{Barker2018}
Jon Barker, Shinji Watanabe, Emmanuel Vincent, and Jan Trmal,
\newblock ``The fifth 'chime' speech separation and recognition challenge:
  Dataset, task and baselines,''
\newblock in {\em Interspeech 2018}, 2018, pp. 1561--1565.

\bibitem{du2018theustc}
Jun Du et~al.,
\newblock ``The ustc-iflytek systems for chime-5 challenge,''
\newblock in {\em CHiME-5}, 2018.

\bibitem{kanda2018hitachi}
Naoyuki Kanda et~al.,
\newblock ``The hitachi/jhu chime-5 system: Advances in speech recognition for
  everyday home environments using multiple microphone arrays,''
\newblock in {\em CHiME-5}, 2018.

\bibitem{fiscus1997post}
Jonathan~G Fiscus,
\newblock ``A post-processing system to yield reduced word error rates:
  Recognizer output voting error reduction (rover),''
\newblock in {\em ASRU 1997}. IEEE, 1997, pp. 347--354.

\bibitem{wang2018stream}
Xiaofei Wang, Ruizhi Li, and Hynek Hermansky,
\newblock ``Stream attention for distributed multi-microphone speech
  recognition,''
\newblock in {\em Interspeech 2018}, 2018, pp. 3033--3037.

\bibitem{xiong2018channel}
Feifei Xiong et~al.,
\newblock ``Channel selection using neural network posterior probability for
  speech recognition with distributed microphone arrays in everyday
  environments,''
\newblock in {\em CHiME-5}, 2018.

\bibitem{mallidi2015uncertainty}
Sri~Harish Mallidi, Tetsuji Ogawa, and Hynek Hermansky,
\newblock ``Uncertainty estimation of dnn classifiers,''
\newblock in {\em ASRU 2015}. IEEE, 2015, pp. 283--288.

\bibitem{chan2016listen}
William Chan, Navdeep Jaitly, Quoc Le, and Oriol Vinyals,
\newblock ``Listen, attend and spell: A neural network for large vocabulary
  conversational speech recognition,''
\newblock in {\em ICASSP 2016}. IEEE, 2016, pp. 4960--4964.

\bibitem{chorowski2015attention}
Jan~K Chorowski, Dzmitry Bahdanau, Dmitriy Serdyuk, Kyunghyun Cho, and Yoshua
  Bengio,
\newblock ``Attention-based models for speech recognition,''
\newblock in {\em NIPS 2015}, 2015, pp. 577--585.

\bibitem{graves2014towards}
Alex Graves and Navdeep Jaitly,
\newblock ``Towards end-to-end speech recognition with recurrent neural
  networks,''
\newblock in {\em ICML 2014}, 2014, pp. 1764--1772.

\bibitem{miao2015eesen}
Yajie Miao, Mohammad Gowayyed, and Florian Metze,
\newblock ``{EESEN}: End-to-end speech recognition using deep {RNN} models and
  {WFST}-based decoding,''
\newblock in {\em ASRU 2015}, 2015, pp. 167--174.

\bibitem{kim2016joint_icassp2017}
Suyoun Kim, Takaaki Hori, and Shinji Watanabe,
\newblock ``Joint {CTC}-attention based end-to-end speech recognition using
  multi-task learning,''
\newblock in {\em ICASSP 2017}, 2017, pp. 4835--4839.

\bibitem{hori2017advances}
Takaaki Hori, Shinji Watanabe, Yu~Zhang, and William Chan,
\newblock ``Advances in joint {CTC}-attention based end-to-end speech
  recognition with a deep {CNN} encoder and {RNN-LM},''
\newblock in {\em Interspeech 2017}, 2017.

\bibitem{watanabe2017hybrid}
Shinji Watanabe, Takaaki Hori, Suyoun Kim, John~R Hershey, and Tomoki Hayashi,
\newblock ``Hybrid ctc/attention architecture for end-to-end speech
  recognition,''
\newblock {\em IEEE Journal of Selected Topics in Signal Processing}, vol. 11,
  no. 8, pp. 1240--1253, 2017.

\bibitem{ochiai2017unified}
Tsubasa Ochiai, Shinji Watanabe, Takaaki Hori, John~R Hershey, and Xiong Xiao,
\newblock ``Unified architecture for multichannel end-to-end speech recognition
  with neural beamforming,''
\newblock {\em IEEE Journal of Selected Topics in Signal Processing}, vol. 11,
  no. 8, pp. 1274--1288, 2017.

\bibitem{braun_2018}
Stefan Braun, Daniel Neil, Jithendar Anumula, Enea Ceolini, and Shih-Chii Liu,
\newblock ``Multi-channel attention for end-to-end speech recognition,''
\newblock in {\em Interspeech 2018}, 2018, pp. 17--21.

\bibitem{ochiai2017multichannel}
Tsubasa Ochiai, Shinji Watanabe, Takaaki Hori, and John~R Hershey,
\newblock ``Multichannel end-to-end speech recognition,''
\newblock {\em arXiv preprint arXiv:1703.04783}, 2017.

\bibitem{kim2017end}
Suyoun Kim, Ian Lane, S.~Kim, and I.~Lane,
\newblock ``End-to-end speech recognition with auditory attention for
  multi-microphone distance speech recognition,''
\newblock in {\em Interspeech 2017}, 2017, pp. 3867--3871.

\bibitem{li2018multi}
Ruizhi Li, Xiaofei Wang, Sri~Harish Mallidi, Takaaki Hori, Shinji Watanabe, and
  Hynek Hermansky,
\newblock ``Multi-encoder multi-resolution framework for end-to-end speech
  recognition,''
\newblock {\em arXiv preprint arXiv:1811.04897}, 2018.

\bibitem{yang2016hierarchical}
Zichao Yang, Diyi Yang, Chris Dyer, Xiaodong He, Alex Smola, and Eduard Hovy,
\newblock ``Hierarchical attention networks for document classification,''
\newblock in {\em NAACL HLT}, 2016, pp. 1480--1489.

\bibitem{hori2017attention}
Chiori Hori, Takaaki Hori, Teng-Yok Lee, Ziming Zhang, Bret Harsham, John~R
  Hershey, Tim~K Marks, and Kazuhiko Sumi,
\newblock ``Attention-based multimodal fusion for video description,''
\newblock in {\em ICCV 2017}. IEEE, 2017, pp. 4203--4212.

\bibitem{libovicky2017attention}
Jind{\v{r}}ich Libovick{\`y} and Jind{\v{r}}ich Helcl,
\newblock ``Attention strategies for multi-source sequence-to-sequence
  learning,''
\newblock in {\em ACL 2017}, 2017, vol.~2, pp. 196--202.

\bibitem{cho2018multilingual}
Jaejin Cho, Murali~Karthick Baskar, Ruizhi Li, Matthew Wiesner, Sri~Harish
  Mallidi, Nelson Yalta, Martin Karafiat, Shinji Watanabe, and Takaaki Hori,
\newblock ``Multilingual sequence-to-sequence speech recognition: architecture,
  transfer learning, and language modeling,''
\newblock in {\em SLT 2018}, 2018.

\bibitem{carletta2005ami}
Jean Carletta, Simone Ashby, Sebastien Bourban, Mike Flynn, Mael Guillemot,
  Thomas Hain, Jaroslav Kadlec, Vasilis Karaiskos, Wessel Kraaij, Melissa
  Kronenthal, et~al.,
\newblock ``The ami meeting corpus: A pre-announcement,''
\newblock in {\em International Workshop on Machine Learning for Multimodal
  Interaction}. Springer, 2005, pp. 28--39.

\bibitem{ravanelli2016realistic}
Mirco Ravanelli, Piergiorgio Svaizer, and Maurizio Omologo,
\newblock ``Realistic multi-microphone data simulation for distant speech
  recognition,''
\newblock in {\em Interspeech 2016}, 2016.

\bibitem{anguera2007acoustic}
Xavier Anguera, Chuck Wooters, and Javier Hernando,
\newblock ``Acoustic beamforming for speaker diarization of meetings,''
\newblock {\em IEEE Transactions on Audio, Speech, and Language Processing},
  vol. 15, no. 7, pp. 2011--2022, 2007.

\bibitem{watanabe2018espnet}
Shinji Watanabe, Takaaki Hori, Shigeki Karita, Tomoki Hayashi, Jiro Nishitoba,
  Yuya Unno, Nelson {Enrique Yalta Soplin}, Jahn Heymann, Matthew Wiesner,
  Nanxin Chen, Adithya Renduchintala, and Tsubasa Ochiai,
\newblock ``Espnet: End-to-end speech processing toolkit,''
\newblock in {\em Interspeech 2018}, 2018, pp. 2207--2211.

\bibitem{hori2018end}
Takaaki Hori, Jaejin Cho, and Shinji Watanabe,
\newblock ``End-to-end speech recognition with word-based {RNN} language
  models,''
\newblock {\em arXiv preprint arXiv:1808.02608}, 2018.

\end{thebibliography}

\end{document}